\documentclass[runningheads]{llncs}
\usepackage{graphicx}
\usepackage{comment}
\usepackage{amsmath,amssymb} 
\usepackage{color}
\usepackage[normalem]{ulem}
\usepackage{float}
\usepackage{subfigure}
\useunder{\uline}{\ul}{}
\newcommand{\etal}{\textit{et al. }}
\usepackage{hyperref}
\hypersetup{colorlinks, linkcolor=red}

\begin{document}
\pagestyle{headings}
\mainmatter
\def\ECCVSubNumber{5170}  

\title{A Metric Learning Reality Check} 


\titlerunning{A Metric Learning Reality Check}
%
\author{Kevin Musgrave\inst{1},
Serge Belongie\inst{1},
Ser-Nam Lim\inst{2}}
\authorrunning{Musgrave et al.}
%
\institute{$^{1}$Cornell Tech \quad $^{2}$Facebook AI}
\maketitle

\begin{abstract}
Deep metric learning papers from the past four years have consistently claimed great advances in accuracy, often more than doubling the performance of decade-old methods. In this paper, we take a closer look at the field to see if this is actually true. We find flaws in the experimental methodology of numerous metric learning papers, and show that the actual improvements over time have been marginal at best. Code is available at \href{https://www.github.com/KevinMusgrave/powerful-benchmarker}{github.com/KevinMusgrave/powerful-benchmarker}.
\end{abstract}

\section{Metric Learning Overview}

\subsection{Why metric learning is important}
Metric learning attempts to map data to an embedding space, where similar data are close together and dissimilar data are far apart. In general, this can be achieved by means of embedding and classification losses. Embedding losses operate on the relationships between samples in a batch, while classification losses include a weight matrix that transforms the embedding space into a vector of class logits.

In cases where a classification loss is applicable, why are embeddings used during test time, instead of the logits or the subsequent softmax values? Typically, embeddings are preferred when the task is some variant of information retrieval, where the goal is to return data that is most similar to a query. An example of this is image search, where the input is a query image, and the output is the most visually similar images in a database. Open-set classification is a variant of this, where test set and training set classes are disjoint. In this situation, query data can be classified based on a nearest neighbors vote, or verified based on distance thresholding in the embedding space. Some notable applications of this are face verification \cite{schroff2015facenet}, and person re-identification \cite{hermans2017defense}. Both have seen improvements in accuracy, largely due to the use of convnets, but also due to loss functions that encourage well-clustered embedding spaces. 

Then there are cases where using a classification loss is not possible. For example, when constructing a dataset, it might be difficult or costly to assign class labels to each sample, and it might be easier to specify the relative similarities between samples in the form of pair or triplet relationships \cite{wilber2014HIT}. Pairs and triplets can also provide additional training signals for existing datasets \cite{Cui2016Bootstrap}. In both cases, there are no explicit labels, so embedding losses become a suitable choice. 

Recently, there has been significant interest in self-supervised learning. This is a form of unsupervised learning where pseudo-labels are applied to the data during training, often via clever use of data augmentations or signals from multiple modalities \cite{owens2018audio,he2019momentum,chen2020simple}. In this case, the pseudo-labels exist to indicate the similarities between data in a particular batch, and as such, they do not have any meaning across training iterations. Thus, embedding losses are favored over classification losses. 

Other applications of embedding losses include learning 3D point cloud features \cite{choy2019fully}, dimensionality reduction for visualization \cite{amid2019trimap}, imitation learning \cite{sermanet2018time}, sequence prediction \cite{oord2018representation}, and even vanilla image classification \cite{khosla2020supervised}.

In the computer vision domain, deep convnets have resulted in dramatic improvements in nearly every subfield, including classification \cite{krizhevsky2012imagenet,he2016deep}, segmentation \cite{long2015fully}, object detection \cite{ren2015faster}, and generative models \cite{goodfellow2014generative}. It is no surprise, then, that deep networks have had a similar effect on metric learning. The combination of the two is often called deep metric learning, and this will be the focus of the remainder of the paper. The rest of this section will briefly review the recent advances in deep metric learning, as well as related work, and the contributions of this paper.  

\subsection{Embedding losses}
Pair and triplet losses provide the foundation for two fundamental approaches to metric learning. A classic pair based method is the contrastive loss \cite{hadsell2006dimensionality}, which attempts to make the distance between positive pairs ($d_{p}$) smaller than some threshold ($m_{pos}$), and the distance between negative pairs ($d_{n}$) larger than some threshold ($m_{neg}$):
\begin{align}
L_{contrastive} & = [d_{p} - m_{pos}]_{+} + [m_{neg} - d_{n}]_{+} \;
\end{align}
(Note that in many implementations, $m_{pos}$ is set to 0.) The theoretical downside of this method is that the same distance threshold is applied to all pairs, even though there may be a large variance in their similarities and dissimilarities. 

The triplet margin loss \cite{weinberger2006distance} theoretically addresses this issue. A triplet consists of an anchor, positive, and negative sample, where the anchor is more similar to the positive than the negative. The triplet margin loss attempts to make the anchor-positive distances ($d_{ap}$) smaller than the anchor-negative distances ($d_{an}$), by a predefined margin ($m$):

\begin{align}
L_{triplet} & = [d_{ap} - d_{an} + m]_{+} \;
\end{align}
This theoretically places fewer restrictions on the embedding space, and allows the model to account for variance in interclass dissimilarities.

A wide variety of losses has since been built on these fundamental concepts. For example, the angular loss \cite{wang2017deep} is a triplet loss where the margin is based on the angles formed by the triplet vectors. The margin loss \cite{wu2017sampling} modifies the contrastive loss by setting $m_{pos} = \beta - \alpha$ and $m_{neg} = \beta + \alpha$, where $\alpha$ is fixed, and $\beta$ is learnable via gradient descent. More recently, Yuan \etal \cite{yuan2019signal} proposed a variation of the contrastive loss based on signal to noise ratios, where each embedding vector is considered signal, and the difference between it and other vectors is considered noise. Other pair losses are based on the softmax function and LogSumExp, which is a smooth approximation of the maximum function. Specifically, the lifted structure loss \cite{oh2016deep} is the contrastive loss but with LogSumExp applied to all negative pairs. The N-Pairs loss \cite{sohn2016improved} applies the softmax function to each positive pair relative to all other pairs. (The N-Pairs loss is also known as InfoNCE \cite{oord2018representation} and NT-Xent \cite{chen2020simple}.) The recent multi similarity loss \cite{wang2019multi} applies LogSumExp to all pairs, but is specially formulated to give weight to different relative similarities among each embedding and its neighbors. The tuplet margin loss \cite{Yu_2019_ICCV} combines LogSumExp with an implicit pair weighting method, while the circle loss \cite{sun2020circle} weights each pair's similarity by its deviation from a pre-determined optimal similarity value. In contrast with these pair and triplet losses, FastAP \cite{cakir2019deep} attempts to optimize for average precision within each batch, using a soft histogram binning technique.


\subsection{Classification losses}
Classification losses are based on the inclusion of a weight matrix, where each column corresponds to a particular class. In most cases, training consists of matrix multiplying the weights with embedding vectors to obtain logits, and then applying a loss function to the logits. The most straightforward case is the normalized softmax loss \cite{wang2017normface,liu2017sphereface,zhai2018classification}, which is identical to cross entropy, but with the columns of the weight matrix L2 normalized. ProxyNCA \cite{movshovitz2017no} is a variation of this, where cross entropy is applied to the Euclidean distances, rather than the cosine similarities, between embeddings and the weight matrix. A number of face verification losses have modified the cross entropy loss with angular margins in the softmax expression. Specifically, SphereFace \cite{liu2017sphereface}, CosFace \cite{wang2018additive,wang2018cosface}, and ArcFace \cite{deng2019arcface} apply multiplicative-angular, additive-cosine, and additive-angular margins, respectively. (It is interesting to note that metric learning papers have consistently left out face verification losses from their experiments, even though there is nothing face-specific about them.) The SoftTriple loss \cite{qian2019softtriple} takes a different approach, by expanding the weight matrix to have multiple columns per class, theoretically providing more flexibility for modeling class variances.

\subsection{Pair and triplet mining}
Mining is the process of finding the best pairs or triplets to train on. There are two broad approaches to mining: offline and online. Offline mining is performed before batch construction, so that each batch is made to contain the most informative samples. This might be accomplished by storing lists of hard negatives \cite{smirnov2017doppelganger}, doing a nearest neighbors search before each epoch \cite{harwood2017smart}, or before each iteration \cite{suh2019stochastic}. In contrast, online mining finds hard pairs or triplets within each randomly sampled batch. Using all possible pairs or triplets is an alternative, but this has two weaknesses: practically, it can consume a lot of memory, and theoretically, it has the tendency to include a large number of easy negatives and positives, causing performance to  plateau quickly. Thus, one intuitive strategy is to select only the most difficult positive and negative samples \cite{hermans2017defense}, but this has been found to produce noisy gradients and convergence to bad local optima \cite{wu2017sampling}. A possible remedy is semihard negative mining, which finds the negative samples in a batch that are close to the anchor, but still further away than the corresponding positive samples \cite{schroff2015facenet}. On the other hand, Wu \etal \cite{wu2017sampling} found that semihard mining makes little progress as the number of semihard negatives drops. They claim that distance-weighted sampling results in a variety of negatives (easy, semihard, and hard), and improved performance. Online mining can also be integrated into the structure of models. Specifically, the hard-aware deeply cascaded method \cite{yuan2017hard} uses models of varying complexity, in which the loss for the complex models only considers the pairs that the simpler models find difficult. Recently, Wang \etal \cite{wang2019multi} proposed a simple pair mining strategy, where negatives are chosen if they are closer to an anchor than its hardest positive, and positives are chosen if they are further from an anchor than its hardest negative.

\subsection{Advanced training methods}
To obtain higher accuracy, many recent papers have gone beyond loss functions or mining techniques. For example, several recent methods incorporate generator networks in their training procedure. Lin \etal \cite{lin2018deep} use a generator as part of their framework for modeling class centers and intraclass variance. Duan \etal \cite{duan2018deep} use a hard-negative generator to expose the model to difficult negatives that might be absent from the training set. Zheng \etal \cite{zheng2019hardness} follow up on this work by using an adaptive interpolation method that creates negatives of varying difficulty, based on the strength of the model. Other sophisticated training methods include HTL \cite{ge2018deep}, ABE \cite{kim2018attention}, MIC \cite{roth2019mic}, and DCES \cite{sanakoyeu2019divide}. HTL constructs a hierarchical class tree at regular intervals during training, to estimate the optimal per-class margin in the triplet margin loss. ABE is an attention based ensemble, where each model learns a different set of attention masks. MIC uses a combination of clustering and encoder networks to disentangle class specific properties from shared characteristics like color and pose. DCES uses a divide and conquer approach, by partitioning the embedding space, and training an embedding layer for each partition separately. 

\subsection{Related work}
Exposing hype and methodological flaws is not new. Papers of this type have been written for machine learning \cite{lipton2018troubling}, image classification \cite{chatfield2014return}, neural network pruning \cite{blalock2020state}, information retrieval \cite{yang2019critically}, recommender systems \cite{dacrema2019we}, and generative adversarial networks \cite{lucic2018gans}. Recently, Fehervari \etal \cite{fehervari2019unbiased} addressed the problem of unfair comparisons in metric learning papers, by evaluating loss functions on a more level playing field. However, they focused mainly on methods from 2017 or earlier, and did not address the issue of hyperparameter tuning on the test set. Concurrent with our work is Roth \etal \cite{roth2020revisiting}, which addresses many of the same flaws that we find, and does an extensive analysis of various loss functions. But again, they do not address the problem of training with test set feedback, and their hyperparameters are tuned using a small grid search around values proposed in the original papers. In contrast, we use cross-validation and bayesian optimization to tune hyperparameters. We find that this significantly minimizes the performance differences between loss functions. See section \ref{ProposedEvaluationMethod} for a complete explanation of our experimental methodology. 

\subsection{Contributions of this paper}
In the following sections, we examine flaws in the current literature, including the problem of unfair comparisons, the weaknesses of commonly used accuracy metrics, and the bad practice of training with test set feedback. We propose a training and evaluation protocol that addresses these flaws, and then run experiments on a variety of loss functions. Our results show that when hyperparameters are properly tuned via cross-validation, most methods perform similarly to one another. This opens up research questions regarding the relationship between hyperparameters and datasets, and the factors limiting open-set accuracy that may be inherent to particular dataset/architecture combinations. As well, by comparing algorithms using proper machine learning practices and a level playing field, the performance gains in future research will better reflect reality, and will be more likely to generalize to other high-impact fields like self-supervised learning.

\section{Flaws in the existing literature}

\subsection{Unfair comparisons}
In order to claim that a new algorithm outperforms existing methods, it’s important to keep as many parameters constant as possible. That way, we can be certain that it was the new algorithm that boosted performance, and not one of the extraneous parameters. This has not been the case with metric learning papers.

One of the easiest ways to improve accuracy is to upgrade the network architecture, yet this fundamental parameter has not been kept constant across papers. Some use GoogleNet, while others use BN-Inception, sometimes referred to as ``Inception with Batch Normalization.” Choice of architecture is important in metric learning, because the networks are typically pretrained on ImageNet, and then finetuned on smaller datasets. Thus, the initial accuracy on the smaller datasets varies depending on the chosen network. One widely-cited paper from 2017 used ResNet50, and then claimed huge performance gains. This is questionable, because the competing methods used GoogleNet, which has significantly lower initial accuracies (see Table \ref{pretrained_accuracy}). Therefore, much of the performance gain likely came from the choice of network architecture, and not their proposed method. In addition, papers have changed the dimensionality of the embedding space, and increasing dimensionality leads to increased accuracy. Therefore, varying this parameter further complicates the task of comparing algorithms.

\bgroup
\def\arraystretch{1.1}
\begin{table}[]
\begin{center}
\caption{\textbf{Recall@1 of models pretrained on ImageNet.} Output embedding sizes were reduced to 512 using PCA and L2 normalized. For each image, the smaller side was scaled to 256, followed by a center-crop to 227x227.}
\label{pretrained_accuracy}
\resizebox{0.6\textwidth}{!}{
\begin{tabular}{l|lll}
                       & CUB200  & Cars196 & SOP   \\ \hline
GoogleNet              & 41.1   & 33.9  & 45.2          \\
BN-Inception           & 51.1   & 46.9  & 50.7          \\
ResNet50               & 48.7   & 43.5  & 52.9          \\
\end{tabular}}
\end{center}
\end{table}
\egroup

Another easy way to improve accuracy is to use more sophisticated image augmentations. In fact, image augmentation strategies have been central to several recent advances in supervised and self-supervised learning \cite{cubuk2019autoaugment,tan2019efficientnet,he2019momentum,chen2020simple}. In the metric learning field, most papers claim to apply the following transformations: resize the image to 256 x 256, randomly crop to 227 x 227, and do a horizontal flip with 50\% chance. But the official open-source implementations of some recent papers show that they are actually using the more sophisticated cropping method described in the original GoogleNet paper. This method randomly changes the location, size, and aspect ratio of each crop, which provides more variability in the training data, and helps combat overfitting.

Papers have also been inconsistent in their choice of optimizer (SGD, Adam, RMSprop etc) and learning rate. The effect on test set accuracy is less clear in this case, as adaptive optimizers like Adam and RMSprop will converge faster, while SGD may lead to better generalization \cite{luo2018adaptive}. Regardless, varying the optimizer and learning rate makes it difficult to do apples-to-apples comparisons.

It is also possible for papers to omit small details that have a big effect on accuracy. For example, in the official open-source code for a 2019 paper, the pretrained ImageNet model has its BatchNorm parameters frozen during training. This can help reduce overfitting, and the authors explain in the code that it results in a 2 point performance boost on the CUB200 dataset. Yet this is not mentioned in their paper.

Finally, most papers do not present confidence intervals for their results, and improvements in accuracy over previous methods often range in the low single digits. Those small improvements would be more meaningful if the results were averaged over multiple runs, and confidence intervals were included.


\subsection{Weakness of commonly used accuracy metrics}
\begin{figure}
\centering
\subfigure{\label{accuracy_toy_example:1}\includegraphics[trim={2cm 9cm 2cm 9cm},clip,width=0.3\textwidth]{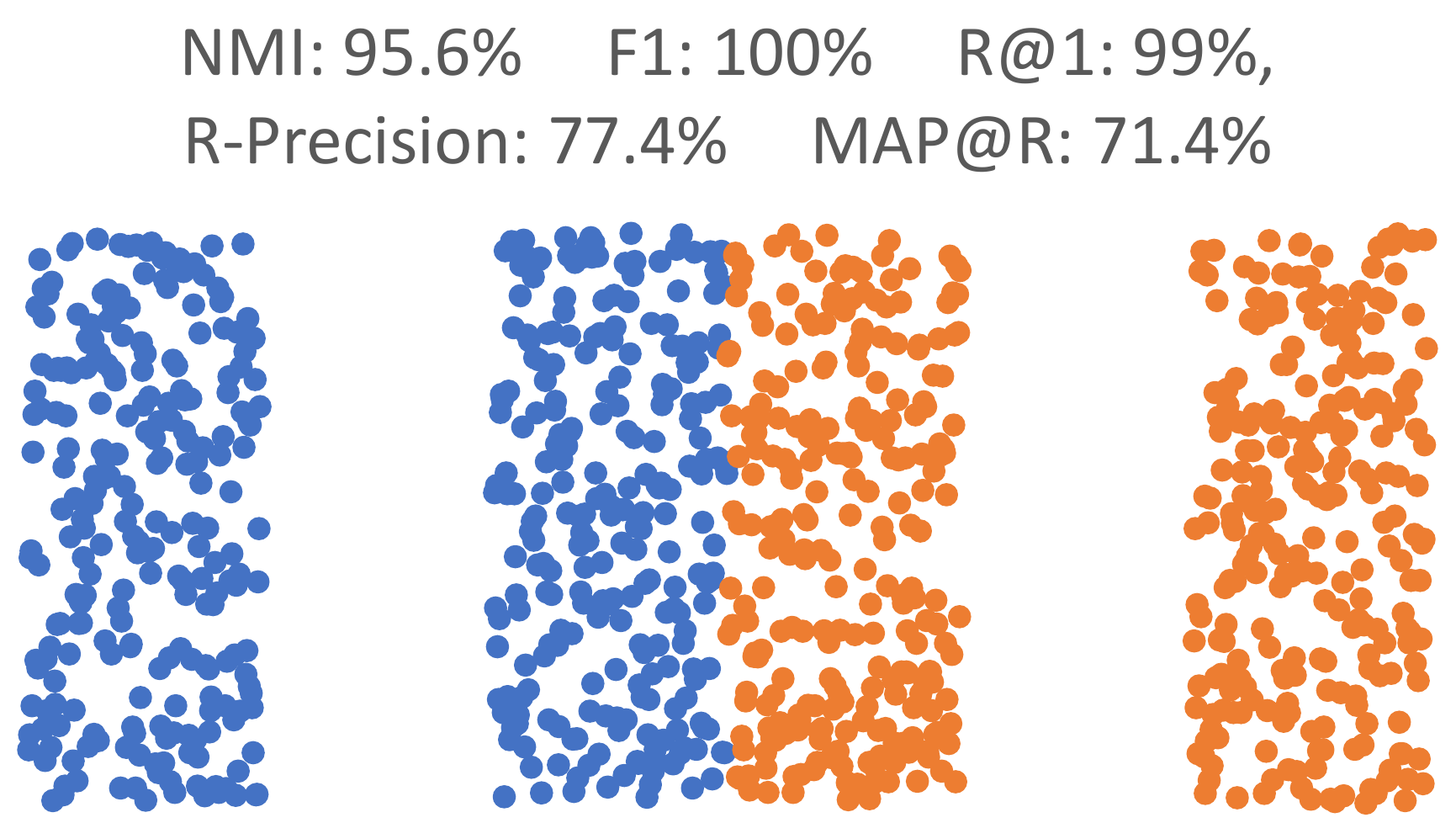}}
\hfill
\subfigure{\label{accuracy_toy_example:2}\includegraphics[trim={2cm 9cm 2cm 9cm},clip,width=0.3\textwidth]{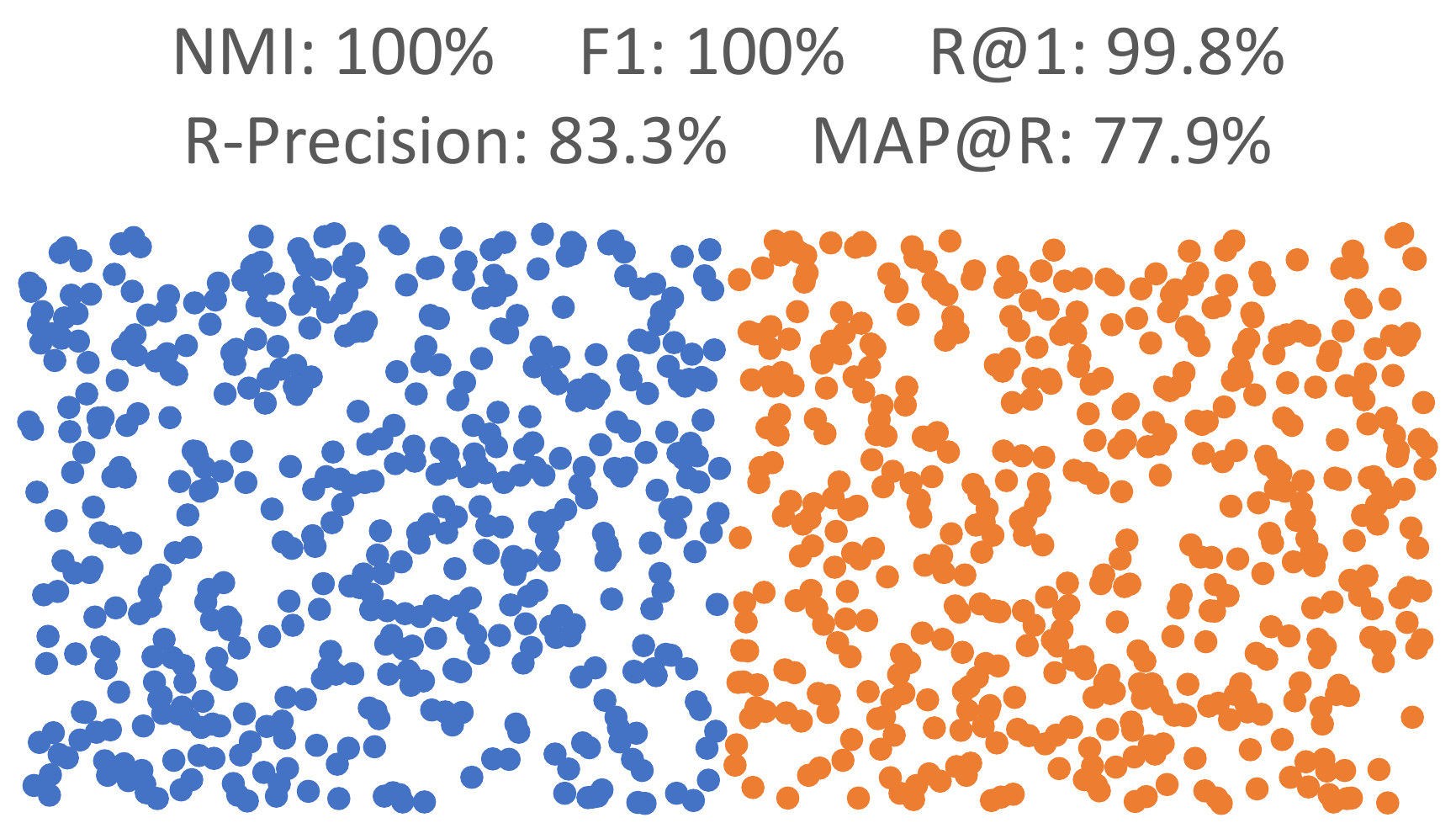}}
\hfill
\subfigure{\label{accuracy_toy_example:3}\includegraphics[trim={2cm 9cm 2cm 9cm},clip,width=0.3\textwidth]{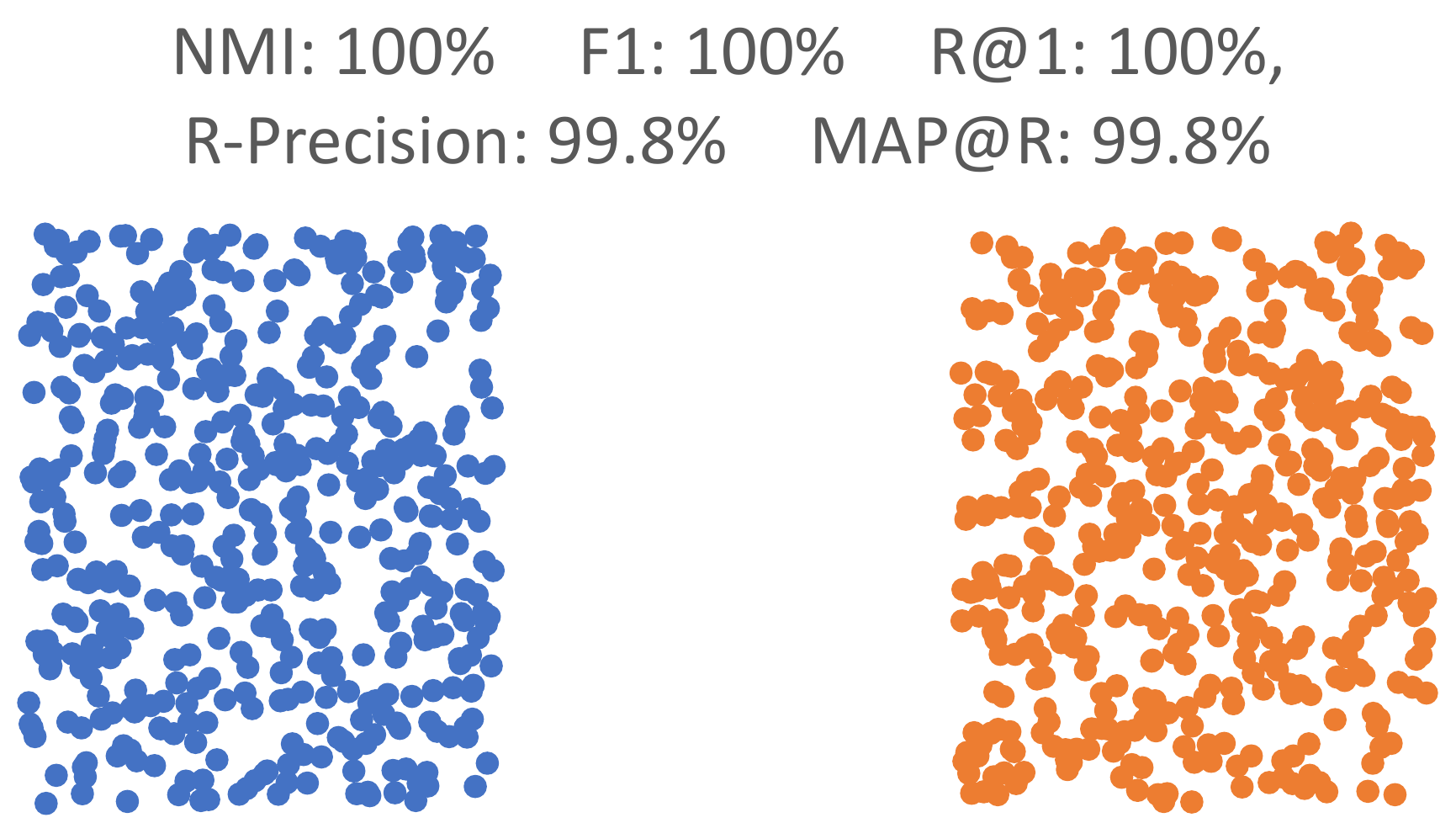}}
\caption{How different accuracy metrics score on three toy examples.}
\label{accuracy_toy_example}
\end{figure}

To report accuracy, most metric learning papers use Recall@K, Normalized Mutual Information (NMI), and the F1 score. But are these necessarily the best metrics to use? Figure \ref{accuracy_toy_example} shows three embedding spaces, and each one scores nearly 100\% Recall@1, even though they have different characteristics. (Note that 100\% Recall@1 means that Recall@K for any K\textgreater1 is also 100\%.) More importantly, Figure \ref{accuracy_toy_example:3} shows a better separation of the classes than Figure \ref{accuracy_toy_example:1}, yet they receive approximately the same score. F1 and NMI also return roughly equal scores for all three embedding spaces. Moreover, they require the embeddings to be clustered, which introduces two factors of variability: the choice of clustering algorithm, and the sensitivity of clustering results to seed initialization. Since we know the ground-truth number of clusters, k-means clustering is the obvious choice and is what is typically used. However, as Figure \ref{accuracy_toy_example} shows, this results in uninformative NMI and F1 scores. Other clustering algorithms could be considered, but each one has its own drawbacks and subtleties. Introducing a clustering algorithm into the evaluation process is simply adding a layer of complexity between the researcher and the embedding space. Instead, we would like an accuracy metric that operates directly on the embedding space, like Recall@K, but that provides more nuanced information.

NMI also tends to give high scores to datasets that have many classes, regardless of the model's true accuracy (see Table \ref{random_initialization_example}). Adjusted Mutual Information \cite{vinh2010information} removes this flaw, but still requires clustering to be done first.

\bgroup
\def\arraystretch{1.1}
\begin{table}[]
\begin{center}
\caption{NMI of embeddings from randomly initialized convnets. CUB200 and Cars196 have about 200 classes, while SOP has about 20,000.}
\label{random_initialization_example}
\resizebox{0.6\textwidth}{!}{
\begin{tabular}{l|lll}
                       & CUB200  & Cars196  &  SOP   \\ \hline
GoogleNet             & 23.6  & 19.1  & 81.2         \\
BN-Inception           & 18.5  & 13.7  & 73.1         \\
ResNet50               & 21.3  & 16.7  & 80.8         \\
\end{tabular}}
\end{center}
\end{table}
\egroup

\subsection{Training with test set feedback}
The majority of papers split each dataset so that the first 50\% of classes are used for the training set, and the remainder are used for the test set. Then during training, the test set accuracy of the model is checked at regular intervals, and the best test set accuracy is reported. In other words, there is no validation set, and model selection and hyperparameter tuning are done with direct feedback from the test set. Some papers do not check performance at regular intervals, and instead report accuracy after training for a predetermined number of iterations. In this case, it is unclear how the number of iterations is chosen, and hyperparameters are still tuned based on test set performance. This breaks one of the most basic commandments of machine learning. Training with test set feedback
leads to overfitting on the test set, and therefore brings into question the steady rise in accuracy over time, as presented in metric learning papers.
\section{Proposed evaluation method}\label{ProposedEvaluationMethod}

The following is an explanation of our experimental methodology, which fixes the flaws described in the previous section.

\subsection{Fair comparisons and reproducibility}
All experiments are run using PyTorch \cite{paszke2019pytorch} with the following settings:
\begin{itemize}
    \item{The trunk model is an ImageNet \cite{russakovsky2015imagenet} pretrained BN-Inception network \cite{ioffe2015batch}, with output embedding size of 128. BatchNorm parameters are frozen during training, to reduce overfitting.}
    \item{The batch size is set to 32. Batches are constructed by first randomly sampling $C$ classes, and then randomly sampling $M$ images for each of the $C$ classes. We set $C=8$ and $M=4$ for embedding losses, and $C=32$ and $M=1$ for classification losses.}
    \item{During training, images are augmented using the random resized cropping strategy. Specifically, we first resize each image so that its shorter side has length 256, then make a random crop that has a size between 40 and 256, and aspect ratio between 3/4 and 4/3. This crop is then resized to 227x227, and flipped horizontally with 50\% probability. During evaluation, images are resized to 256 and then center cropped to 227.}
    \item{All network parameters are optimized using RMSprop with learning rate 1e-6. We chose RMSprop because it converges faster than SGD, and seems to generalize better than Adam, based on a small set of experiments. For loss functions that include their own learnable weights (e.g. ArcFace), we use RMSprop but leave the learning rate as a hyperparameter to be optimized.}
    \item{Embeddings are L2 normalized before computing the loss, and during evaluation.}
\end{itemize}
Source code, configuration files, and other supplementary material are available at \href{https://www.github.com/KevinMusgrave/powerful-benchmarker}{github.com/KevinMusgrave/powerful-benchmarker}.

\subsection{Informative accuracy metrics}
We measure accuracy using Mean Average Precision at R (MAP@R), which combines the ideas of Mean Average Precision and R-precision.
\begin{itemize}
    \item{R-Precision is defined as follows: For each query\footnote{A query is an image for which we are trying to find similar images, and the references are the searchable database.}, let $R$ be the total number of references that are the same class as the query. Find the $R$ nearest references to the query, and let $r$ be the number of those nearest references that are the same class as the query. The score for the query is $\frac{r}{R}$.}
    \item{One weakness of R-precision is that it does not account for the ranking of the correct retrievals. So we instead use MAP@R, which is Mean Average Precision with the number of nearest neighbors for each sample set to R. For a single query:}
    \begin{equation}
    \text{MAP@R} = \frac{1}{R}\sum_{i=1}^{R}P(i)
    \end{equation}
    \begin{equation}
    P(i) = 
    \begin{cases}
    \text{precision at } i,& \text{if the ith retrieval is correct}\\
    0,                      & \text{otherwise}
    \end{cases}
    \end{equation}
\end{itemize}

\noindent The benefits of MAP@R are that it is more informative than Recall@1 (see Figure \ref{accuracy_toy_example} and Table \ref{HypotheticalSearchResults}), it can be computed directly from the embedding space (no clustering step required), it is easy to understand, and it rewards well clustered embedding spaces. MAP@R is also more stable than Recall@1. Across our experiments, we computed the lag-one autocorrelation of the validation accuracy during training: Recall@1 = 0.73 and MAP@R = 0.81. Thus, MAP@R is less noisy, making it easier to select the best performing model checkpoints.

\bgroup
\def\arraystretch{1.1}
\begin{table}[]
\begin{center}
\caption{Accuracy metrics on hypothetical retrieval results. The accuracy numbers represent percentages. Assume R = 10. Despite the clear differences, Recall@1 scores all four retrieval results at 100\%, so it fails to capture important information. }
\label{HypotheticalSearchResults}
\resizebox{\textwidth}{!}{
\begin{tabular}{l|c|c|c}
\textbf{Retrieval results} & \textbf{Recall@1} & \textbf{R-Precision}   & \textbf{MAP@R} \\ \hline
10 results, of which only the 1st is correct & 100 & 10 & 10 \\ \hline
10 results, of which the 1st and 10th are correct & 100 & 20 & 12 \\ \hline
10 results, of which the 1st and 2nd are correct & 100 & 20 & 20 \\ \hline
10 results, of which all 10 are correct & 100 & 100 & 100 \\ \hline
\end{tabular}}
\end{center}
\end{table}
\egroup

\noindent In our results tables in section \ref{experiments_section}, we present R-precision and MAP@R. For the sake of comparisons to previous papers, we also show Precision@1 (also known as ``Recall@1" in the previous sections and in metric learning papers).

\subsection{Hyperparameter search via cross validation}
To find the best loss function hyperparameters, we run 50 iterations of bayesian optimization. Each iteration consists of 4-fold cross validation:
\begin{itemize}
\item {The first half of classes are used for cross validation, and the 4 partitions are created deterministically: the first 0-12.5\% of classes make up the first partition, the next 12.5-25\% of classes make up the second partition, and so on. The training set comprises 3 of the 4 partitions, and cycles through all leave-one-out possibilities. As a result, the training and validation sets are always class-disjoint, so optimizing for validation set performance should be a good proxy for accuracy on open-set tasks. Training stops when validation accuracy plateaus.}
\item {The second half of classes are used as the test set. This is the same setting that metric learning papers have used for years, and we use it so that results can be compared more easily to past papers.}
\end{itemize}

\noindent Hyperparameters are optimized to maximize the average validation accuracy. For the best hyperparameters, the highest-accuracy checkpoint for each training set partition is loaded, and its embeddings for the test set are computed and L2 normalized. Then we compute accuracy using two methods:
    \begin{enumerate}
        \item{\textbf{Concatenated (512-dim)}: For each sample in the test set, we concatenate the 128-dim embeddings of the 4 models to get 512-dim embeddings, and then L2 normalize. We then report the accuracy of these embeddings.}
        \item{\textbf{Separated (128-dim)}: For each sample in the test set, we compute the accuracy of the 128-dim embeddings separately, and therefore obtain 4 different accuracies, one for each model's embeddings. We then report the average of these accuracies.}
    \end{enumerate}
    
\noindent We do 10 training runs using the best hyperparameters, and report the average across these runs, as well as confidence intervals. This way our results are less subject to random seed noise.

\section{Experiments}\label{experiments_section}
\subsection{Losses and datasets}
We ran experiments on 13 losses, and 1 loss+miner combination, and prioritized methods from recent conferences (see Table \ref{ExperimentLossesOverview}). For every loss, we used the settings described in section \ref{ProposedEvaluationMethod}, and we ran experiments on three widely used metric learning datasets: CUB200 \cite{WahCUB_200_2011}, Cars196 \cite{KrauseStarkDengFei-Fei_3DRR2013}, and Stanford Online Products (SOP) \cite{oh2016deep}. We chose these datasets because they have been the standard for several years, and we wanted our results to be easily comparable to prior papers. Tables \ref{CUB200Results}-\ref{SOPResults} show the mean accuracy across 10 training runs, as well as the 95\% confidence intervals where applicable. Bold represents the best mean accuracy. We also include the accuracy of the pretrained model, the embeddings of which are reduced to 512 or 128, using PCA. In the supplementary material, we show results for CUB200 using a batch size of 256 instead of 32. The results are roughly the same, with the exception of FastAP, which gets a significant boost in accuracy, and performs on par with the rest of the methods. 

\bgroup
\def\arraystretch{1.1}
\begin{table}[]
\begin{center}
\caption{Accuracy on CUB200}
\label{CUB200Results}
\resizebox{\textwidth}{!}{
\begin{tabular}{c|ccc|ccc}
          & \multicolumn{3}{c|}{\textbf{Concatenated (512-dim)}}  & \multicolumn{3}{c}{\textbf{Separated (128-dim)}}  \\ \hline
                 & \textbf{P@1} & \textbf{RP} & \textbf{MAP@R}   & \textbf{P@1} & \textbf{RP} & \textbf{MAP@R} \\ \hline
        Pretrained & 51.05 & 24.85 & 14.21 & 50.54 & 25.12 & 14.53 \\ \hline

        Contrastive  & $\textbf{68.13} \pm \textbf{0.31}$  & $37.24 \pm 0.28$  & $26.53  \pm 0.29$ & $59.73 \pm 0.40$ & $31.98 \pm 0.29$ & $21.18 \pm	0.28$ \\ \hline
        Triplet  & $64.24 \pm 0.26$	& $34.55 \pm 0.24$ & $23.69 \pm	0.23$ & $55.76 \pm 0.27$ & $29.55 \pm 0.16$ & $18.75 \pm 0.15$  \\ \hline
        NT-Xent    & $66.61 \pm 0.29$ & $35.96 \pm 0.21$ & $25.09 \pm 0.22$ & $58.12 \pm	0.23$ & $30.81 \pm 0.17$ & $19.87 \pm 0.16$ \\ \hline
        ProxyNCA  & $65.69 \pm 0.43$ & $35.14 \pm 0.26$ & $24.21 \pm 0.27$ & $57.88 \pm 0.30$ & $30.16 \pm	0.22$ & $19.32 \pm 0.21$  \\ \hline
        Margin  & $63.60 \pm 0.48$ &	$33.94 \pm	0.27$ &	$23.09 \pm 	0.27$ & $54.78 \pm 0.30$ &	$28.86 \pm 0.18$ & $18.11 \pm 0.17$    \\ \hline
        Margin/class  & $64.37 \pm 0.18$ & $34.59 \pm 0.16$ & $23.71 \pm 0.16$ & $55.56 \pm	0.16$ &	$29.32 \pm 0.15$ & $18.51 \pm	0.13$ \\ \hline
        N. Softmax &  $ 65.65 \pm 0.30$ & $35.99 \pm 0.15$ & $25.25 \pm 0.13$ & $58.75 \pm 0.19$ & $31.75 \pm	0.12$ &	$20.96 \pm	0.11$ \\ \hline
        CosFace  & $67.32 \pm	0.32$ & $\textbf{37.49} \pm	\textbf{0.21}$ & $\textbf{26.70} \pm \textbf{0.23}$ & $59.63 \pm 0.36$ & $31.99 \pm 0.22$	& $21.21 \pm 0.22$ \\ \hline
        ArcFace  & $67.50 \pm 0.25$ & $37.31 \pm 0.21$ & $26.45 \pm 0.20$ & $\textbf{60.17} \pm \textbf{0.32}$ & $\textbf{32.37} \pm \textbf{0.17}$ & $\textbf{21.49} \pm \textbf{0.16}$ \\ \hline
        FastAP  & $63.17 \pm 0.34$ & $34.20 \pm 0.20$ &	$23.53 \pm 0.20$ & $55.58 \pm 0.31$ & $29.72 \pm 0.16$ & $19.09 \pm 0.16$ \\ \hline
        SNR & $66.44 \pm 0.56$ & $36.56 \pm 0.34$ & $25.75 \pm 0.36$ & $58.06 \pm 0.39$ & $31.21 \pm 0.28$ & $20.43 \pm 0.28$ \\ \hline
        MS  & $65.04 \pm 0.28$ & $35.40 \pm 0.12$ &	$24.70 \pm 0.13$ & $57.60 \pm 0.24$ & $30.84 \pm 0.13$ &	$20.15 \pm 0.14$ \\ \hline
        MS+Miner & $67.73 \pm 0.18$ & $37.37 \pm 0.19$ & $26.52 \pm 0.18$ & $59.41 \pm 0.30$ & $31.93 \pm 0.15$ & $21.01 \pm 0.14$ \\ \hline
        SoftTriple & $67.27 \pm 0.39$ & $37.34 \pm 0.19$ & $26.51 \pm 0.20$ & $59.94 \pm 0.33$ & $32.12 \pm 0.14$ & $21.31 \pm 0.14$ \\ \hline

\end{tabular}}
\end{center}
\end{table}
\egroup

\bgroup
\def\arraystretch{1.1}
\begin{table}[]
\begin{center}
\caption{Accuracy on Cars196}
\label{Cars196Results}
\resizebox{\textwidth}{!}{
\begin{tabular}{c|ccc|ccc}
          & \multicolumn{3}{c|}{\textbf{Concatenated (512-dim)}}  & \multicolumn{3}{c}{\textbf{Separated (128-dim)}}  \\ \hline
                 & \textbf{P@1} & \textbf{RP} & \textbf{MAP@R}   & \textbf{P@1} & \textbf{RP} & \textbf{MAP@R} \\ \hline
        Pretrained & 46.89 & 13.77 & 5.91 & 43.27 & 13.37 & 5.64\\ \hline
        Contrastive & $81.78 \pm 0.43$ & $35.11 \pm 0.45$ & $24.89 \pm 0.50$ & $69.80 \pm 0.38$ & $27.78 \pm 0.34$ & $17.24 \pm	0.35$ \\ \hline        
        Triplet & $79.13 \pm 0.42$ & $33.71 \pm 0.45$ & $23.02 \pm 0.51$ & $65.68 \pm 0.58$ & $26.67 \pm 0.36$ & $15.82 \pm 0.36$ \\ \hline 
        NT-Xent & $80.99 \pm 0.54$ &	$34.96 \pm 0.38$ &	$24.40 \pm 0.41$ & $68.16 \pm 0.36$	& $27.66 \pm 0.23$ & $16.78 \pm 0.24$ \\ \hline
        ProxyNCA & $83.56 \pm 0.27$	& $35.62 \pm 0.28$ & $25.38 \pm 0.31$ & $73.46 \pm 0.23$ & $28.90 \pm 0.22$ & $18.29 \pm 0.22$ \\ \hline  
        Margin  & $81.16 \pm 0.50$ & $34.82 \pm 0.31$ & $24.21 \pm 0.34$ & $68.24 \pm 0.35$	& $27.25 \pm 0.19$ & $16.40 \pm 0.20$ \\ \hline 
        Margin / class & $80.04 \pm 0.61$ &	$33.78 \pm 0.51$ & $23.11 \pm 0.55$ & $67.54 \pm 0.60$ & $26.68 \pm 0.40$ &	$15.88 \pm 0.39$ \\ \hline 
        N. Softmax & $83.16 \pm 0.25$ &	$36.20 \pm 0.26$ & $26.00 \pm 0.30$ & $72.55 \pm 0.18$ &$29.35 \pm 0.20$ & $18.73 \pm 0.20$ \\ \hline  
        CosFace & $\textbf{85.52} \pm \textbf{0.24}$ & $37.32 \pm 0.28$ & $27.57 \pm 0.30$ & $\textbf{74.67} \pm \textbf{0.20}$	& $29.01 \pm 0.11$ & $18.80 \pm 0.12$ \\ \hline         
        ArcFace & $85.44 \pm 0.28$ &	$37.02 \pm 0.29$ &	$27.22 \pm 0.30$ & $72.10 \pm 0.37$ &	$27.29 \pm 0.17$ &	$17.11 \pm 0.18$ \\ \hline  
        FastAP & $78.45 \pm 0.52$ & $33.61 \pm 0.54$ & $23.14 \pm 0.56$ & $65.08 \pm 0.36$ & $26.59 \pm 0.36$ & $15.94 \pm 0.34$ \\ \hline  
        SNR & $82.02 \pm 0.48$ &	$35.22 \pm 0.43$ &	$25.03 \pm 0.48$ & $69.69 \pm 0.46$ & $27.55 \pm 0.25$ & $17.13 \pm 0.26$ \\ \hline         
        MS & $85.14 \pm 0.29$ & $\textbf{38.09} \pm \textbf{0.19}$ & $\textbf{28.07} \pm \textbf{0.22}$ & $73.77 \pm 0.19$ & $\textbf{29.92} \pm \textbf{0.16}$ & $\textbf{19.32} \pm \textbf{0.18}$ \\ \hline  
        MS+Miner & $83.67 \pm 0.34$ & $37.08 \pm 0.31$ & $27.01 \pm 0.35$ & $71.80 \pm 0.22$ & $29.44 \pm 0.21$ & $18.86 \pm 0.20$ \\ \hline 
        SoftTriple & $84.49 \pm 0.26$ & $37.03 \pm 0.21$ & $27.08 \pm 0.21$ & $73.69 \pm 0.21$ & $29.29 \pm 0.16$ & $18.89 \pm 0.16$ \\ \hline  
\end{tabular}}
\end{center}
\end{table}
\egroup

\bgroup
\def\arraystretch{1.1}
\begin{table}[]
\begin{center}
\caption{Accuracy on SOP}
\label{SOPResults}
\resizebox{\textwidth}{!}{
\begin{tabular}{c|ccc|ccc}
          & \multicolumn{3}{c|}{\textbf{Concatenated (512-dim)}}  & \multicolumn{3}{c}{\textbf{Separated (128-dim)}}  \\ \hline
                 & \textbf{P@1} & \textbf{RP} & \textbf{MAP@R}   & \textbf{P@1} & \textbf{RP} & \textbf{MAP@R} \\ \hline
         Pretrained & 50.71 & 25.97 & 23.44 & 47.25 & 23.84 & 21.36 \\ \hline
        Contrastive & $73.12 \pm 0.20$ & $47.29 \pm 0.24$ & $44.39 \pm 0.24$ & $69.34 \pm 0.26$ & $43.41 \pm 0.28$ & $40.37 \pm 0.28$ \\ \hline    
        Triplet & $72.65 \pm 0.28$ & $46.46 \pm 0.38$ & $43.37 \pm 0.37$ & $67.33 \pm 0.34$ & $40.94 \pm 0.39$ & $37.70 \pm 0.38$ \\ \hline
        NT-Xent & $74.22 \pm 0.22$ & $48.35 \pm 0.26$ & $45.31 \pm 0.25$ & $69.88 \pm 0.19$ & $43.51 \pm 0.21$ & $40.31 \pm 0.20$ \\ \hline
        ProxyNCA & $75.89 \pm 0.17$ & $50.10 \pm 0.22$ & $47.22 \pm 0.21$ & $71.30 \pm 0.20$ & $44.71 \pm 0.21$ & $41.74 \pm 0.21$  \\ \hline       
        Margin & $70.99 \pm 0.36$ & $44.94 \pm 0.43$ & $41.82 \pm 0.43$ & $65.78 \pm 0.34$ & $39.71 \pm 0.40$ & $36.47 \pm 0.39$ \\ \hline
        Margin / class & $72.36 \pm 0.30$ & $46.41 \pm 0.40$ & $43.32 \pm 0.41$ & $67.56 \pm 0.42$ &	$41.37 \pm 0.48$ & $38.15 \pm 0.49$ \\ \hline
        N. Softmax & $75.67 \pm 0.17$ & $50.01 \pm 0.22$ & $47.13 \pm 0.22$ & $\textbf{71.65} \pm \textbf{0.14}$ & $\textbf{45.32} \pm \textbf{0.17}$ & $\textbf{42.35} \pm \textbf{0.16}$ \\ \hline       
        CosFace & $75.79 \pm 0.14$ & $49.77 \pm 0.19$ & $46.92 \pm 0.19$ & $70.71 \pm 0.19$ & $43.56 \pm 0.21$ &	$40.69 \pm 0.21$ \\ \hline
        ArcFace  & $\textbf{76.20} \pm \textbf{0.27}$ & $\textbf{50.27} \pm \textbf{0.38}$ & $\textbf{47.41} \pm \textbf{0.40}$ & $70.88 \pm 1.51$ & $44.00 \pm 1.26$ & $41.11 \pm 1.22$ \\ \hline
        FastAP & $72.59 \pm 0.26$ & $46.60 \pm 0.29$ & $43.57 \pm 0.28$ & $68.13 \pm 0.25$ & $42.06 \pm 0.25$ & $38.88 \pm 0.25$ \\ \hline
        SNR & $73.40 \pm 0.09$ & $47.43 \pm 0.13$ & $44.54 \pm 0.13$ & $69.45 \pm 0.10$ &	$43.34 \pm 0.12$ & $40.31 \pm 0.12$ \\ \hline
        MS  & $74.50 \pm 0.24$ & $48.77 \pm 0.32$ & $45.79 \pm 0.32$ & $70.43 \pm 0.33$ & $44.25 \pm 0.38$ & $41.15 \pm 0.38$ \\ \hline
        MS+Miner & $75.09 \pm 0.17$ & $49.51 \pm 0.20$ &	$46.55 \pm 0.20$ & $71.25 \pm 0.15$ & $45.19 \pm 0.16$ & $42.10 \pm 0.16$ \\ \hline
        SoftTriple & $76.12 \pm 0.17$ & $50.21 \pm 0.18$ & $47.35 \pm 0.19$ & $70.88 \pm 0.20$ & $43.83 \pm 0.20$ & $40.92 \pm 0.20$ \\ \hline

\end{tabular}}
\end{center}
\end{table}
\egroup

\begin{table}[]
\caption{The losses covered in our experiments. Note that NT-Xent is the name we used in our code, but it is also known as N-Pairs or InfoNCE. For the Margin loss, we tested two versions: ``Margin" uses the same $\beta$ value for all training classes, and ``Margin / class" uses a separate $\beta$ for each training class. In both versions, $\beta$ is learned during training. Face verification losses have been consistently left out of metric learning papers, so we included two losses (CosFace and ArcFace) from that domain. (We used only the loss functions from those two papers. We did not train on any face datasets or use any model trained on faces.)}
\label{ExperimentLossesOverview}
\begin{center}
\resizebox{0.65\textwidth}{!}{
\begin{tabular}{l|l|l}
\hline
\textbf{Method}                                    & \textbf{Year} & \textbf{Loss type} \\ \hline
Contrastive \cite{hadsell2006dimensionality}        & 2006          & Embedding          \\
Triplet \cite{weinberger2006distance}                                           & 2006          & Embedding          \\
NT-Xent \cite{sohn2016improved,oord2018representation,chen2020simple} & 2016 & Embedding \\
ProxyNCA \cite{movshovitz2017no}                                          & 2017          & Classification     \\
Margin \cite{wu2017sampling}                                            & 2017          & Embedding          \\
Margin / class \cite{wu2017sampling}                                            & 2017          & Embedding          \\
Normalized Softmax (N. Softmax) \cite{wang2017normface,liu2017sphereface,zhai2018classification} & 2017          & Classification     \\
CosFace \cite{wang2018additive,wang2018cosface}                     & 2018          & Classification     \\
ArcFace \cite{deng2019arcface}                      & 2019          & Classification     \\
FastAP \cite{cakir2019deep}                                            & 2019          & Embedding          \\
Signal to Noise Ratio Contrastive (SNR) \cite{yuan2019signal}                 & 2019          & Embedding          \\
MultiSimilarity (MS) \cite{wang2019multi}                               & 2019          & Embedding          \\
MS+Miner \cite{wang2019multi} & 2019          & Embedding          \\
SoftTriple \cite{qian2019softtriple}                                         & 2019          & Classification     \\ \hline

\end{tabular}}
\end{center}
\end{table}


\subsection{Papers versus reality}

\begin{figure}
\centering
\subfigure[The trend according to papers]{\label{PaperClaimsOverTime:1}\includegraphics[trim={4.2cm 3.5cm 4.5cm 2.2cm},clip,width=0.48\textwidth]{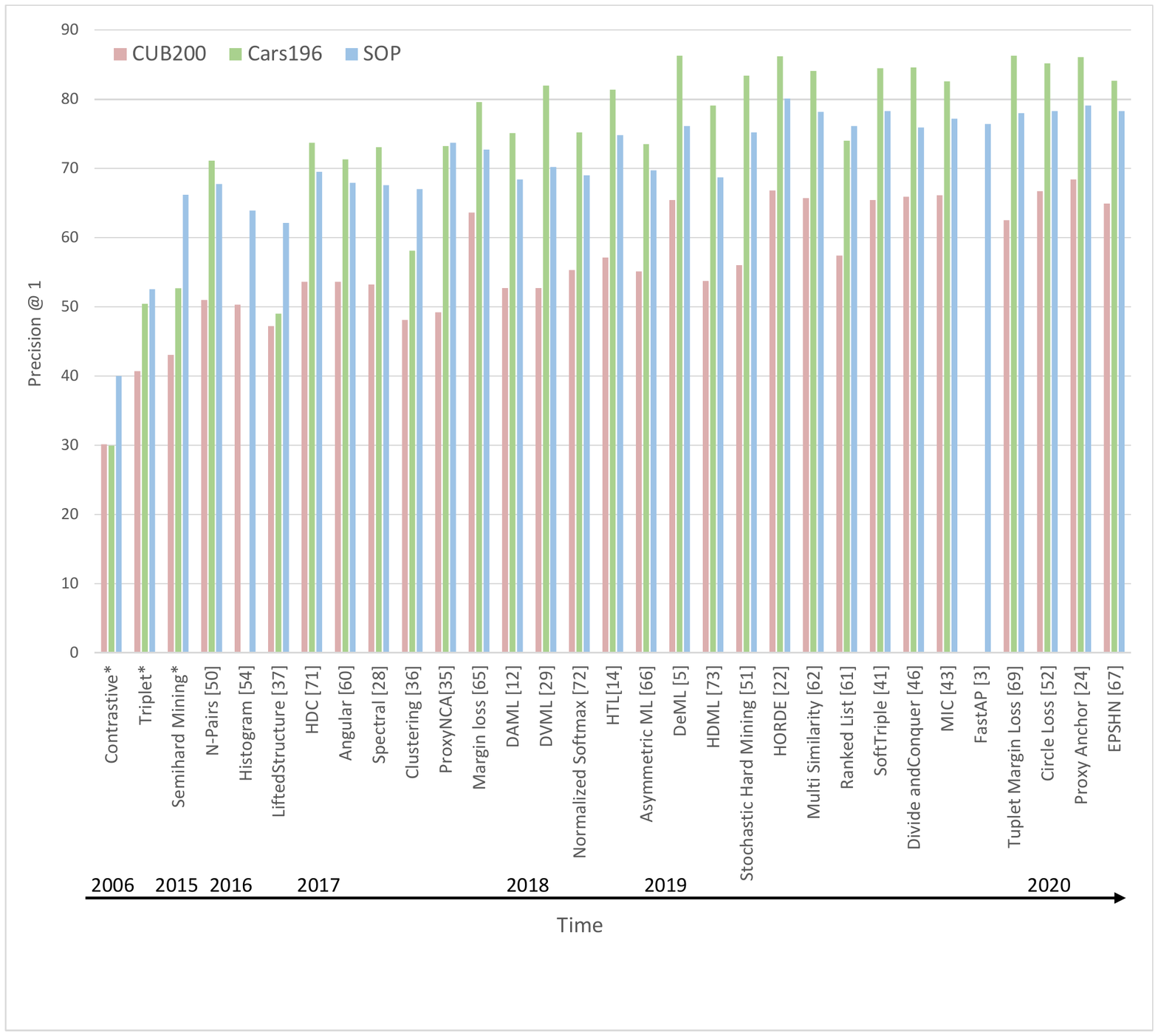}}
\hfill
\subfigure[The trend according to reality]{\label{PaperClaimsOverTime:2}\includegraphics[trim={4.2cm 3.5cm 4.5cm 2.2cm},clip,width=0.48\textwidth]{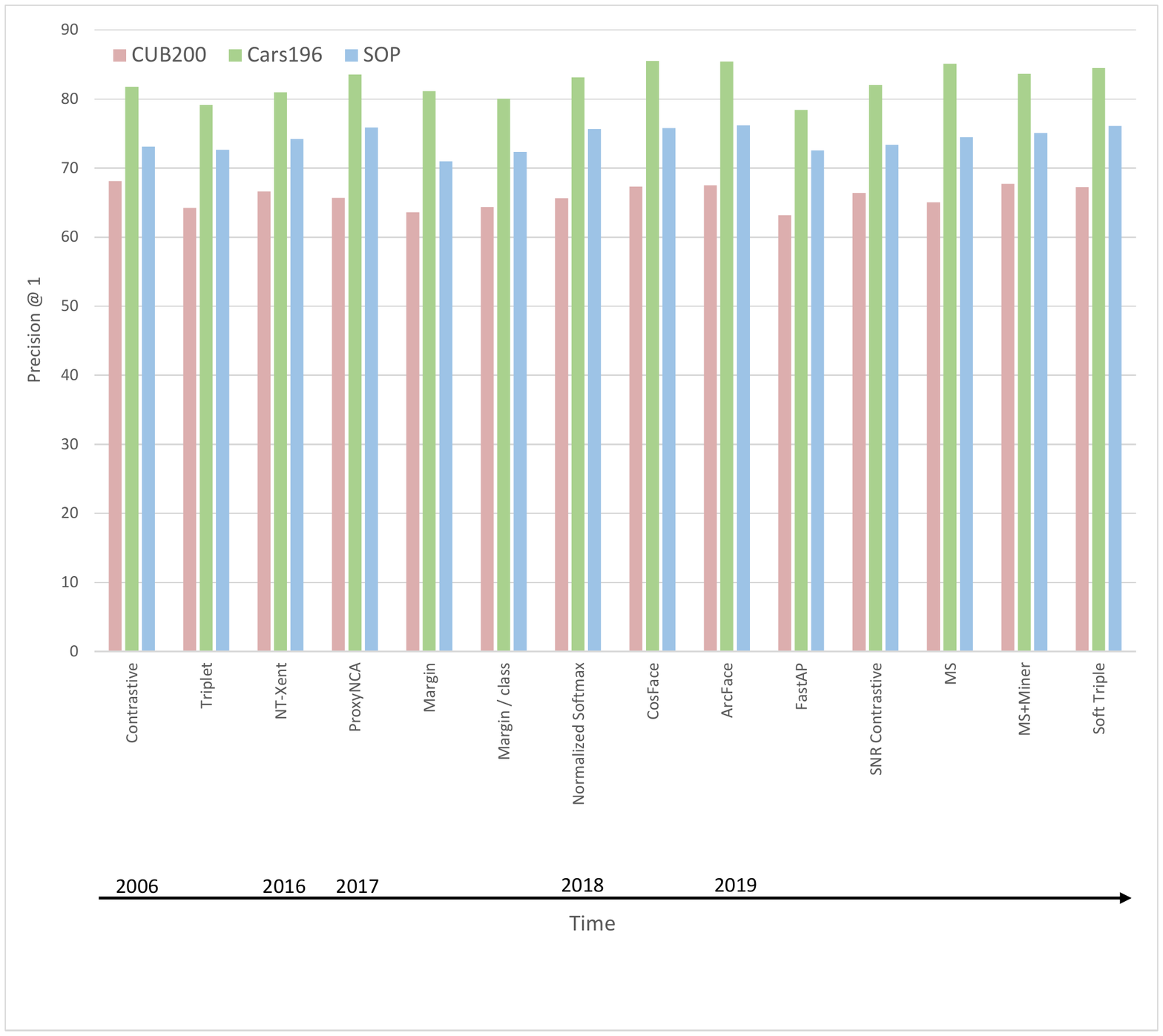}}
\caption{Papers versus Reality: the trend of Precision@1 of various methods over the years. In a), the baseline methods have * next to them, which indicates that their numbers are the average reported accuracy from all papers that included those baselines.}
\label{PaperClaimsOverTime}
\end{figure}

First, let's consider the general trend of paper results. Figure \ref{PaperClaimsOverTime:1} shows the inexorable rise in accuracy we have all come to expect in this field, with modern methods completely obliterating old ones.

\begin{figure}
\centering
\subfigure[Relative improvement over the contrastive loss]{\includegraphics[trim={2cm 3cm 2cm 3cm},clip,width=0.48\textwidth]{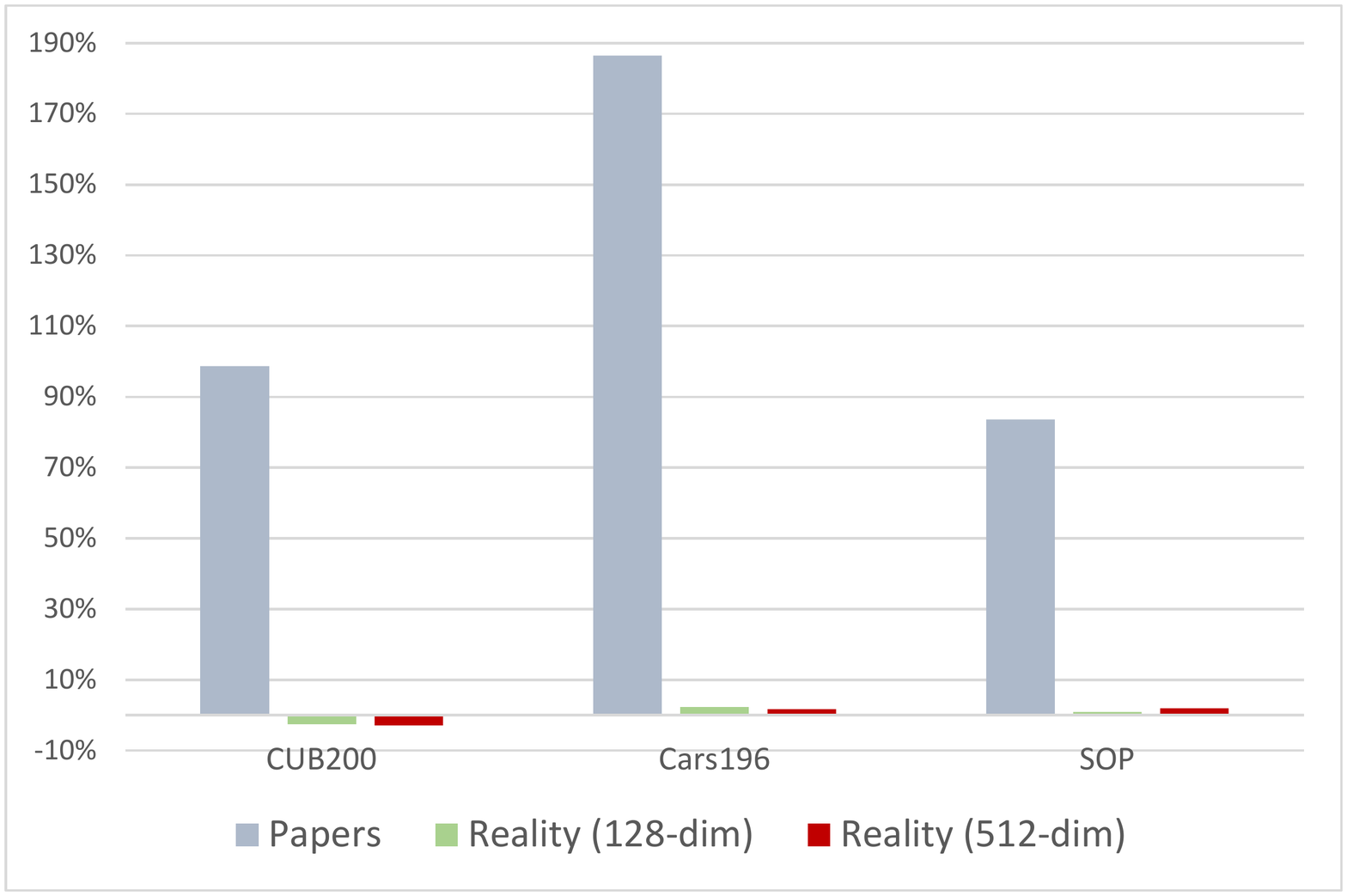}}
\hfill
\subfigure[Relative improvement over the triplet loss]{\includegraphics[trim={2cm 3cm 2cm 3cm},clip,width=0.48\textwidth]{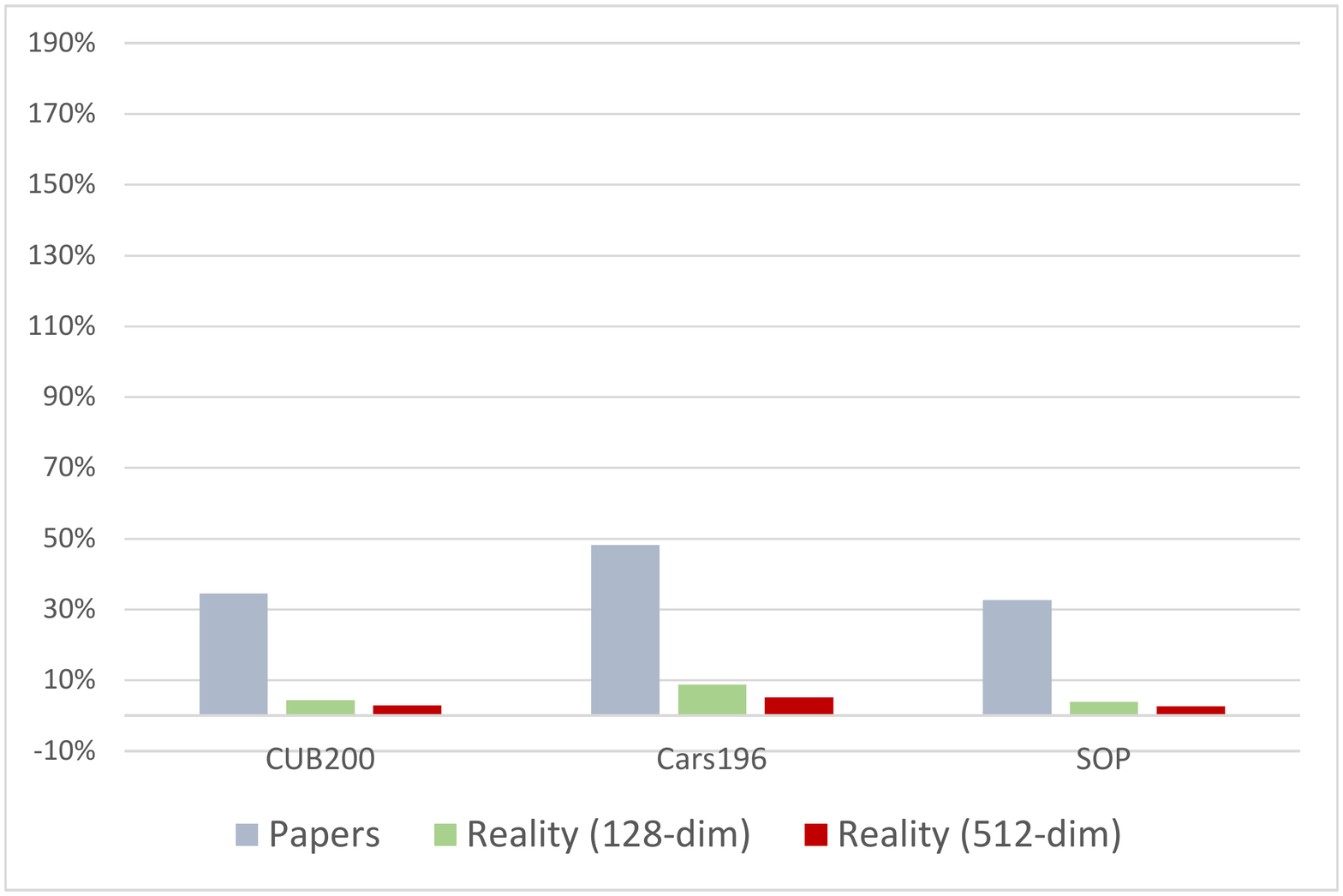}}
\caption{Papers versus Reality: we look at the results tables of all methods presented in Figure \ref{PaperClaimsOverTime:1}. 11 of these include the contrastive loss, and 12 include the triplet loss (without semihard mining). For each paper, we compute the relative percentage improvement of their proposed method over their reported result for the contrastive or triplet loss, and then take the average improvement across papers (grey bars in the above figures). The green and red bars are the average relative improvement that we obtain, in the separated 128-dim and concatenated 512-dim settings, respectively. For the ``reality" numbers in (a) we excluded the FastAP loss from the calculation, since it was a poor performing outlier in our experiments, and we excluded the triplet loss since we consider it a baseline method. Likewise for the ``reality" numbers in (b), we excluded the FastAP and contrastive losses from the calculation.}
\label{PapersVsReality}
\end{figure}

But how do the claims made in papers stack up against reality? We find that papers have drastically overstated improvements over the two classic methods, the contrastive and triplet loss (see Figure \ref{PapersVsReality}). For example, many papers show relative improvements exceeding 100\% when compared with the contrastive loss, and nearing 50\% when compared with the triplet loss. This arises because of the extremely low accuracies that are attributed to these losses. Some of these numbers seem to originate from the 2016 paper on the lifted structure loss \cite{oh2016deep}. In their implementation of the contrastive and triplet loss, they sample $N/2$ pairs and $N/3$ triplets per batch, where $N$ is the batch size. Thus, they utilize only a tiny fraction of the total information provided in each batch. Furthermore, they set the triplet margin to 1, whereas the optimal value tends to be around 0.1. Despite these implementation flaws, most papers simply keep citing the low numbers instead of trying to obtain a more reasonable baseline by implementing the losses themselves. 

With good implementations of those baseline losses, a level playing field, and proper machine learning practices, we obtain the trend as shown in Figure \ref{PaperClaimsOverTime:2}. The trend appears to be a relatively flat line, indicating that the methods perform similarly to one another, whether they were introduced in 2006 or 2019. In other words, metric learning algorithms have not made the spectacular progress that they claim to have made. This brings into question the results of other cutting edge papers not covered in our experiments. It also raises doubts about the value of the hand-wavy theoretical explanations in metric learning papers. If a paper attempts to explain the performance gains of its proposed method, and it turns out that those performance gains are non-existent, then their explanation must be invalid as well.

\section{Conclusion}
In this paper, we uncovered several flaws in the current metric learning literature, namely:

\begin{itemize}
\item{Unfair comparisons caused by changes in network architecture, embedding size, image augmentation method, and optimizers.}
\item{The use of accuracy metrics that are either misleading, or do not a provide a complete picture of the embedding space.}
\item{Training without a validation set, i.e. with test set feedback.}
\end{itemize}

We then ran experiments with these issues fixed, and found that state of the art loss functions perform marginally better than, and sometimes on par with, classic methods. This is in stark contrast with the claims made in papers, in which accuracy has risen dramatically over time. 

Future work could explore the relationship between optimal hyperparameters and dataset/architecture combinations, as well as the reasons for why different losses are performing similarly to one another. Of course, pushing the state-of-the-art in accuracy is another research direction. If proper machine learning practices are followed, and comparisons to prior work are done in a fair manner, the results of future metric learning papers will better reflect reality, and will be more likely to generalize to other high-impact areas like self-supervised learning.

\section{Acknowledgements}
This work is supported by a Facebook AI research grant awarded to Cornell University.

\clearpage
%
%
\bibliographystyle{splncs04}
\bibliography{egbib}

\begin{thebibliography}{10}
\providecommand{\url}[1]{\texttt{#1}}
\providecommand{\urlprefix}{URL }
\providecommand{\doi}[1]{https://doi.org/#1}

\bibitem{amid2019trimap}
Amid, E., Warmuth, M.K.: Trimap: Large-scale dimensionality reduction using
  triplets. arXiv preprint arXiv:1910.00204  (2019)

\bibitem{blalock2020state}
Blalock, D., Ortiz, J.J.G., Frankle, J., Guttag, J.: What is the state of
  neural network pruning? arXiv preprint arXiv:2003.03033  (2020)

\bibitem{cakir2019deep}
Cakir, F., He, K., Xia, X., Kulis, B., Sclaroff, S.: Deep metric learning to
  rank. In: Proceedings of the IEEE Conference on Computer Vision and Pattern
  Recognition. pp. 1861--1870 (2019)

\bibitem{chatfield2014return}
Chatfield, K., Simonyan, K., Vedaldi, A., Zisserman, A.: Return of the devil in
  the details: Delving deep into convolutional nets. arXiv preprint
  arXiv:1405.3531  (2014)

\bibitem{chen2019hybrid}
Chen, B., Deng, W.: Hybrid-attention based decoupled metric learning for
  zero-shot image retrieval. In: Proceedings of the IEEE Conference on Computer
  Vision and Pattern Recognition. pp. 2750--2759 (2019)

\bibitem{chen2020simple}
Chen, T., Kornblith, S., Norouzi, M., Hinton, G.: A simple framework for
  contrastive learning of visual representations. arXiv preprint
  arXiv:2002.05709  (2020)

\bibitem{choy2019fully}
Choy, C., Park, J., Koltun, V.: Fully convolutional geometric features. In:
  Proceedings of the IEEE International Conference on Computer Vision. pp.
  8958--8966 (2019)

\bibitem{cubuk2019autoaugment}
Cubuk, E.D., Zoph, B., Mane, D., Vasudevan, V., Le, Q.V.: Autoaugment: Learning
  augmentation strategies from data. In: Proceedings of the IEEE conference on
  computer vision and pattern recognition. pp. 113--123 (2019)

\bibitem{Cui2016Bootstrap}
Cui, Y., Zhou, F., Lin, Y., Belongie, S.: Fine-grained categorization and
  dataset bootstrapping using deep metric learning with humans in the loop. In:
  Computer Vision and Pattern Recognition (CVPR). Las Vegas, NV (2016),
  \url{http://vision.cornell.edu/se3/wp-content/uploads/2016/04/1950.pdf}

\bibitem{dacrema2019we}
Dacrema, M.F., Cremonesi, P., Jannach, D.: Are we really making much progress?
  a worrying analysis of recent neural recommendation approaches. In:
  Proceedings of the 13th ACM Conference on Recommender Systems. pp. 101--109
  (2019)

\bibitem{deng2019arcface}
Deng, J., Guo, J., Xue, N., Zafeiriou, S.: Arcface: Additive angular margin
  loss for deep face recognition. In: Proceedings of the IEEE Conference on
  Computer Vision and Pattern Recognition. pp. 4690--4699 (2019)

\bibitem{duan2018deep}
Duan, Y., Zheng, W., Lin, X., Lu, J., Zhou, J.: Deep adversarial metric
  learning. In: Proceedings of the IEEE Conference on Computer Vision and
  Pattern Recognition. pp. 2780--2789 (2018)

\bibitem{fehervari2019unbiased}
Fehervari, I., Ravichandran, A., Appalaraju, S.: Unbiased evaluation of deep
  metric learning algorithms. arXiv preprint arXiv:1911.12528  (2019)

\bibitem{ge2018deep}
Ge, W.: Deep metric learning with hierarchical triplet loss. In: Proceedings of
  the European Conference on Computer Vision (ECCV). pp. 269--285 (2018)

\bibitem{goodfellow2014generative}
Goodfellow, I., Pouget-Abadie, J., Mirza, M., Xu, B., Warde-Farley, D., Ozair,
  S., Courville, A., Bengio, Y.: Generative adversarial nets. In: Advances in
  neural information processing systems. pp. 2672--2680 (2014)

\bibitem{hadsell2006dimensionality}
Hadsell, R., Chopra, S., LeCun, Y.: Dimensionality reduction by learning an
  invariant mapping. In: 2006 IEEE Computer Society Conference on Computer
  Vision and Pattern Recognition (CVPR'06). vol.~2, pp. 1735--1742. IEEE (2006)

\bibitem{harwood2017smart}
Harwood, B., Kumar, B., Carneiro, G., Reid, I., Drummond, T., et~al.: Smart
  mining for deep metric learning. In: Proceedings of the IEEE International
  Conference on Computer Vision. pp. 2821--2829 (2017)

\bibitem{he2019momentum}
He, K., Fan, H., Wu, Y., Xie, S., Girshick, R.: Momentum contrast for
  unsupervised visual representation learning. arXiv preprint arXiv:1911.05722
  (2019)

\bibitem{he2016deep}
He, K., Zhang, X., Ren, S., Sun, J.: Deep residual learning for image
  recognition. In: Proceedings of the IEEE conference on computer vision and
  pattern recognition. pp. 770--778 (2016)

\bibitem{hermans2017defense}
Hermans, A., Beyer, L., Leibe, B.: In defense of the triplet loss for person
  re-identification. arXiv preprint arXiv:1703.07737  (2017)

\bibitem{ioffe2015batch}
Ioffe, S., Szegedy, C.: Batch normalization: Accelerating deep network training
  by reducing internal covariate shift. In: International Conference on Machine
  Learning. pp. 448--456 (2015)

\bibitem{jacob2019metric}
Jacob, P., Picard, D., Histace, A., Klein, E.: Metric learning with horde:
  High-order regularizer for deep embeddings. In: Proceedings of the IEEE
  International Conference on Computer Vision. pp. 6539--6548 (2019)

\bibitem{khosla2020supervised}
Khosla, P., Teterwak, P., Wang, C., Sarna, A., Tian, Y., Isola, P., Maschinot,
  A., Liu, C., Krishnan, D.: Supervised contrastive learning. arXiv preprint
  arXiv:2004.11362  (2020)

\bibitem{kim2020proxy}
Kim, S., Kim, D., Cho, M., Kwak, S.: Proxy anchor loss for deep metric
  learning. In: Proceedings of the IEEE/CVF Conference on Computer Vision and
  Pattern Recognition. pp. 3238--3247 (2020)

\bibitem{kim2018attention}
Kim, W., Goyal, B., Chawla, K., Lee, J., Kwon, K.: Attention-based ensemble for
  deep metric learning. In: Proceedings of the European Conference on Computer
  Vision (ECCV). pp. 736--751 (2018)

\bibitem{KrauseStarkDengFei-Fei_3DRR2013}
Krause, J., Stark, M., Deng, J., Fei-Fei, L.: 3d object representations for
  fine-grained categorization. In: 4th International IEEE Workshop on 3D
  Representation and Recognition (3dRR-13). Sydney, Australia (2013)

\bibitem{krizhevsky2012imagenet}
Krizhevsky, A., Sutskever, I., Hinton, G.E.: Imagenet classification with deep
  convolutional neural networks. In: Advances in neural information processing
  systems. pp. 1097--1105 (2012)

\bibitem{law2017deep}
Law, M.T., Urtasun, R., Zemel, R.S.: Deep spectral clustering learning. In:
  International Conference on Machine Learning. pp. 1985--1994 (2017)

\bibitem{lin2018deep}
Lin, X., Duan, Y., Dong, Q., Lu, J., Zhou, J.: Deep variational metric
  learning. In: Proceedings of the European Conference on Computer Vision
  (ECCV). pp. 689--704 (2018)

\bibitem{lipton2018troubling}
Lipton, Z.C., Steinhardt, J.: Troubling trends in machine learning scholarship.
  arXiv preprint arXiv:1807.03341  (2018)

\bibitem{liu2017sphereface}
Liu, W., Wen, Y., Yu, Z., Li, M., Raj, B., Song, L.: Sphereface: Deep
  hypersphere embedding for face recognition. In: Proceedings of the IEEE
  conference on computer vision and pattern recognition. pp. 212--220 (2017)

\bibitem{long2015fully}
Long, J., Shelhamer, E., Darrell, T.: Fully convolutional networks for semantic
  segmentation. In: Proceedings of the IEEE conference on computer vision and
  pattern recognition. pp. 3431--3440 (2015)

\bibitem{lucic2018gans}
Lucic, M., Kurach, K., Michalski, M., Gelly, S., Bousquet, O.: Are gans created
  equal? a large-scale study. In: Advances in neural information processing
  systems. pp. 700--709 (2018)

\bibitem{luo2018adaptive}
Luo, L., Xiong, Y., Liu, Y.: Adaptive gradient methods with dynamic bound of
  learning rate. In: International Conference on Learning Representations
  (2019), \url{https://openreview.net/forum?id=Bkg3g2R9FX}

\bibitem{movshovitz2017no}
Movshovitz-Attias, Y., Toshev, A., Leung, T.K., Ioffe, S., Singh, S.: No fuss
  distance metric learning using proxies. In: Proceedings of the IEEE
  International Conference on Computer Vision. pp. 360--368 (2017)

\bibitem{oh2017deep}
Oh~Song, H., Jegelka, S., Rathod, V., Murphy, K.: Deep metric learning via
  facility location. In: Proceedings of the IEEE Conference on Computer Vision
  and Pattern Recognition. pp. 5382--5390 (2017)

\bibitem{oh2016deep}
Oh~Song, H., Xiang, Y., Jegelka, S., Savarese, S.: Deep metric learning via
  lifted structured feature embedding. In: Proceedings of the IEEE Conference
  on Computer Vision and Pattern Recognition. pp. 4004--4012 (2016)

\bibitem{oord2018representation}
Oord, A.v.d., Li, Y., Vinyals, O.: Representation learning with contrastive
  predictive coding. arXiv preprint arXiv:1807.03748  (2018)

\bibitem{owens2018audio}
Owens, A., Efros, A.A.: Audio-visual scene analysis with self-supervised
  multisensory features. European Conference on Computer Vision (ECCV)  (2018)

\bibitem{paszke2019pytorch}
Paszke, A., Gross, S., Massa, F., Lerer, A., Bradbury, J., Chanan, G., Killeen,
  T., Lin, Z., Gimelshein, N., Antiga, L., et~al.: Pytorch: An imperative
  style, high-performance deep learning library. In: Advances in Neural
  Information Processing Systems. pp. 8024--8035 (2019)

\bibitem{qian2019softtriple}
Qian, Q., Shang, L., Sun, B., Hu, J., Li, H., Jin, R.: Softtriple loss: Deep
  metric learning without triplet sampling. In: Proceedings of the IEEE
  International Conference on Computer Vision. pp. 6450--6458 (2019)

\bibitem{ren2015faster}
Ren, S., He, K., Girshick, R., Sun, J.: Faster r-cnn: Towards real-time object
  detection with region proposal networks. In: Advances in neural information
  processing systems. pp. 91--99 (2015)

\bibitem{roth2019mic}
Roth, K., Brattoli, B., Ommer, B.: Mic: Mining interclass characteristics for
  improved metric learning. In: Proceedings of the IEEE International
  Conference on Computer Vision. pp. 8000--8009 (2019)

\bibitem{roth2020revisiting}
Roth, K., Milbich, T., Sinha, S., Gupta, P., Ommer, B., Cohen, J.P.: Revisiting
  training strategies and generalization performance in deep metric learning
  (2020)

\bibitem{russakovsky2015imagenet}
Russakovsky, O., Deng, J., Su, H., Krause, J., Satheesh, S., Ma, S., Huang, Z.,
  Karpathy, A., Khosla, A., Bernstein, M., et~al.: Imagenet large scale visual
  recognition challenge. International journal of computer vision
  \textbf{115}(3),  211--252 (2015)

\bibitem{sanakoyeu2019divide}
Sanakoyeu, A., Tschernezki, V., Buchler, U., Ommer, B.: Divide and conquer the
  embedding space for metric learning. In: Proceedings of the IEEE Conference
  on Computer Vision and Pattern Recognition. pp. 471--480 (2019)

\bibitem{schroff2015facenet}
Schroff, F., Kalenichenko, D., Philbin, J.: Facenet: A unified embedding for
  face recognition and clustering. In: Proceedings of the IEEE conference on
  computer vision and pattern recognition. pp. 815--823 (2015)

\bibitem{sermanet2018time}
Sermanet, P., Lynch, C., Chebotar, Y., Hsu, J., Jang, E., Schaal, S., Levine,
  S., Brain, G.: Time-contrastive networks: Self-supervised learning from
  video. In: 2018 IEEE International Conference on Robotics and Automation
  (ICRA). pp. 1134--1141. IEEE (2018)

\bibitem{smirnov2017doppelganger}
Smirnov, E., Melnikov, A., Novoselov, S., Luckyanets, E., Lavrentyeva, G.:
  Doppelganger mining for face representation learning. In: Proceedings of the
  IEEE International Conference on Computer Vision Workshops. pp. 1916--1923
  (2017)

\bibitem{sohn2016improved}
Sohn, K.: Improved deep metric learning with multi-class n-pair loss objective.
  In: Advances in Neural Information Processing Systems. pp. 1857--1865 (2016)

\bibitem{suh2019stochastic}
Suh, Y., Han, B., Kim, W., Lee, K.M.: Stochastic class-based hard example
  mining for deep metric learning. In: Proceedings of the IEEE Conference on
  Computer Vision and Pattern Recognition. pp. 7251--7259 (2019)

\bibitem{sun2020circle}
Sun, Y., Cheng, C., Zhang, Y., Zhang, C., Zheng, L., Wang, Z., Wei, Y.: Circle
  loss: A unified perspective of pair similarity optimization. In: Proceedings
  of the IEEE/CVF Conference on Computer Vision and Pattern Recognition. pp.
  6398--6407 (2020)

\bibitem{tan2019efficientnet}
Tan, M., Le, Q.V.: Efficientnet: Rethinking model scaling for convolutional
  neural networks. arXiv preprint arXiv:1905.11946  (2019)

\bibitem{ustinova2016learning}
Ustinova, E., Lempitsky, V.: Learning deep embeddings with histogram loss. In:
  Advances in Neural Information Processing Systems. pp. 4170--4178 (2016)

\bibitem{vinh2010information}
Vinh, N.X., Epps, J., Bailey, J.: Information theoretic measures for
  clusterings comparison: Variants, properties, normalization and correction
  for chance. The Journal of Machine Learning Research  \textbf{11},
  2837--2854 (2010)

\bibitem{WahCUB_200_2011}
Wah, C., Branson, S., Welinder, P., Perona, P., Belongie, S.: {The Caltech-UCSD
  Birds-200-2011 Dataset}. Tech. Rep. CNS-TR-2011-001, California Institute of
  Technology (2011)

\bibitem{wang2018additive}
Wang, F., Cheng, J., Liu, W., Liu, H.: Additive margin softmax for face
  verification. IEEE Signal Processing Letters  \textbf{25}(7),  926--930
  (2018)

\bibitem{wang2017normface}
Wang, F., Xiang, X., Cheng, J., Yuille, A.L.: Normface: L2 hypersphere
  embedding for face verification. In: Proceedings of the 25th ACM
  international conference on Multimedia. pp. 1041--1049 (2017)

\bibitem{wang2018cosface}
Wang, H., Wang, Y., Zhou, Z., Ji, X., Gong, D., Zhou, J., Li, Z., Liu, W.:
  Cosface: Large margin cosine loss for deep face recognition. In: Proceedings
  of the IEEE Conference on Computer Vision and Pattern Recognition. pp.
  5265--5274 (2018)

\bibitem{wang2017deep}
Wang, J., Zhou, F., Wen, S., Liu, X., Lin, Y.: Deep metric learning with
  angular loss. In: Proceedings of the IEEE International Conference on
  Computer Vision. pp. 2593--2601 (2017)

\bibitem{wang2019ranked}
Wang, X., Hua, Y., Kodirov, E., Hu, G., Garnier, R., Robertson, N.M.: Ranked
  list loss for deep metric learning. In: Proceedings of the IEEE Conference on
  Computer Vision and Pattern Recognition. pp. 5207--5216 (2019)

\bibitem{wang2019multi}
Wang, X., Han, X., Huang, W., Dong, D., Scott, M.R.: Multi-similarity loss with
  general pair weighting for deep metric learning. In: Proceedings of the IEEE
  Conference on Computer Vision and Pattern Recognition. pp. 5022--5030 (2019)

\bibitem{weinberger2006distance}
Weinberger, K.Q., Blitzer, J., Saul, L.K.: Distance metric learning for large
  margin nearest neighbor classification. In: Advances in neural information
  processing systems. pp. 1473--1480 (2006)

\bibitem{wilber2014HIT}
Wilber, M., Kwak, S., Belongie, S.: Cost-effective hits for relative similarity
  comparisons. In: Human Computation and Crowdsourcing (HCOMP). Pittsburgh
  (2014), \url{/se3/wp-content/uploads/2015/01/hcomp-conference-paper.pdf,
  http://arxiv.org/abs/1404.3291}

\bibitem{wu2017sampling}
Wu, C.Y., Manmatha, R., Smola, A.J., Krahenbuhl, P.: Sampling matters in deep
  embedding learning. In: Proceedings of the IEEE International Conference on
  Computer Vision. pp. 2840--2848 (2017)

\bibitem{xu2019deep}
Xu, X., Yang, Y., Deng, C., Zheng, F.: Deep asymmetric metric learning via rich
  relationship mining. In: Proceedings of the IEEE Conference on Computer
  Vision and Pattern Recognition. pp. 4076--4085 (2019)

\bibitem{xuan2020improved}
Xuan, H., Stylianou, A., Pless, R.: Improved embeddings with easy positive
  triplet mining. In: The IEEE Winter Conference on Applications of Computer
  Vision. pp. 2474--2482 (2020)

\bibitem{yang2019critically}
Yang, W., Lu, K., Yang, P., Lin, J.: Critically examining the" neural hype"
  weak baselines and the additivity of effectiveness gains from neural ranking
  models. In: Proceedings of the 42nd International ACM SIGIR Conference on
  Research and Development in Information Retrieval. pp. 1129--1132 (2019)

\bibitem{Yu_2019_ICCV}
Yu, B., Tao, D.: Deep metric learning with tuplet margin loss. In: The IEEE
  International Conference on Computer Vision (ICCV) (October 2019)

\bibitem{yuan2019signal}
Yuan, T., Deng, W., Tang, J., Tang, Y., Chen, B.: Signal-to-noise ratio: A
  robust distance metric for deep metric learning. In: Proceedings of the IEEE
  Conference on Computer Vision and Pattern Recognition. pp. 4815--4824 (2019)

\bibitem{yuan2017hard}
Yuan, Y., Yang, K., Zhang, C.: Hard-aware deeply cascaded embedding. In:
  Proceedings of the IEEE international conference on computer vision. pp.
  814--823 (2017)

\bibitem{zhai2018classification}
Zhai, A., Wu, H.Y.: Classification is a strong baseline for deep metric
  learning. arXiv preprint arXiv:1811.12649  (2018)

\bibitem{zheng2019hardness}
Zheng, W., Chen, Z., Lu, J., Zhou, J.: Hardness-aware deep metric learning. In:
  Proceedings of the IEEE Conference on Computer Vision and Pattern
  Recognition. pp. 72--81 (2019)

\end{thebibliography}


\nocite{ustinova2016learning}
\nocite{sohn2016improved}
\nocite{oh2016deep}
\nocite{yuan2017hard}
\nocite{wang2017deep}
\nocite{law2017deep}
\nocite{oh2017deep}
\nocite{movshovitz2017no}
\nocite{wu2017sampling}
\nocite{duan2018deep}
\nocite{lin2018deep}
\nocite{zhai2018classification}
\nocite{ge2018deep}
\nocite{xu2019deep}
\nocite{chen2019hybrid}
\nocite{zheng2019hardness}
\nocite{suh2019stochastic}
\nocite{jacob2019metric}
\nocite{wang2019multi}
\nocite{wang2019ranked}
\nocite{qian2019softtriple}
\nocite{sanakoyeu2019divide}
\nocite{roth2019mic}
\nocite{cakir2019deep}
\nocite{Yu_2019_ICCV}
\nocite{sun2020circle}
\nocite{kim2020proxy}
\nocite{xuan2020improved}

\end{document}